\documentclass[sigconf]{acmart-me}

\usepackage{booktabs} 
\usepackage{url}
\usepackage{color}
\usepackage{enumitem}
\setcopyright{rightsretained}

\acmDOI{}

\acmISBN{}

\acmConference[MediaEval'21]{Multimedia Evaluation Workshop}{December 13-15 2021}{Online} 
\acmYear{2021}
\copyrightyear{}

\acmPrice{}

\begin{document}
\title{Detecting COVID-19 Conspiracy Theories with Transformers and TF-IDF}

\author{Haoming Guo\textsuperscript{1}, Tianyi Huang\textsuperscript{2}, Huixuan Huang\textsuperscript{3},\\ Mingyue Fan\textsuperscript{4}, Gerald Friedland\textsuperscript{5}}
\affiliation{\textsuperscript{12345}University of California, Berkeley}
\email{mike0221@berkeley.edu, 
tianyihuang@berkeley.edu, hhx@berkeley.edu, migofan@berkeley.edu, fractor@berkeley.edu}

\renewcommand{\shortauthors}{H. Guo et al.}
\renewcommand{\shorttitle}{Task name as it appears on https://multimediaeval.github.io/editions/2021/}

\begin{abstract}
The sharing of fake news and conspiracy theories on social media has wide-spread negative effects. By designing and applying different machine learning models, researchers have made progress in detecting fake news from text. However, existing research places a heavy emphasis on general, common-sense fake news, while in reality fake news often involves rapidly changing topics and domain-specific vocabulary. In this paper, we present our methods and results for three fake news detection tasks at MediaEval benchmark 2021 that specifically involve COVID-19 related topics. We experiment with a group of text-based models including Support Vector Machines, Random Forest, BERT, and RoBERTa. We find that a pre-trained transformer yields the best validation results, but a randomly initialized transformer with smart design  can also be trained to reach accuracies close to that of the pre-trained transformer.
\end{abstract}

%
%
%
%
%


\maketitle

\section{Introduction}
\label{sec:intro}
This paper presents several methods in detecting online conspiracy theories from tweets. The task overview paper \cite{pogorelov2021fakenews} \cite{10.1145/3472720.3483617} describes the dataset more in-depth as well as providing information on how the dataset was constructed. In later sections, we describe our methods including support vector machines, random forest, and transformers in depth and present the performance of these models on the provided dataset.

\section{Related Work}
\label{sec:work}
There are a variety of methods towards the goal of detecting text stance and fake news, including false knowledge, writing style, propagation patterns, and source credibility.\cite{Zhou_2020} Much of the textual methodologies focus on writing style, as the information is completely embedded in the textual data. 

FNC-1, a similar benchmark on fake news and stance detection from texts, has received much attention from researchers\cite{fnc}. Using handcrafted features and a Multi-Layer Perceptron model has proved to perform well on the task\cite{fnc}. Recently, transformer architecture was shown to exceed previous results on a wide range of natural language tasks.\cite{vaswani2017attention} Furthermore, Slovikovskaya et al. showed that fine-tuning pretrained transformers achieves state-of-the-art on the FNC-1 benchmark.\cite{fnc-bert} It was also shown that BERT is top-performing on other fake news detection task.\cite{Khan_2021}

\section{Approach}
\label{sec:approach}
In this section, we describe the methodologies behind our feature design, model choice and model training.

\subsection{TF-IDF}
We used Term Frequency Inverse Document Frequency (TF-IDF) to create classifiers based only on the plain tweet texts, which is described in this section. TF-IDF, which is robust to understand the text content, is useful for finding important and related words or phrases in the text. We used the IDF transformer from the scikit-learn framework to get the weight of different words. After removing all the punctuations and stop words, we implemented the TF-IDF Vectorizer to preprocess the training data. 
We proceeded with creating several classifiers based on scikit-learn framework, such as naïve bayes classifier, decision tree classifier, Random Forest (RF) classifier, Support Vector Machine (SVM) classifier, etc. Having a quick look into the training dataset, we found that the dataset is imbalanced, since the number of a certain category was much lesser than others. In order to overcome this, we included the class weight to balance the datasets. During our experiments, we observed RF classifier and SVM classifier perform much better and more efficiently than other methods. Thus, we decided to use SVM classifier and RF classifier for future prediction. Furthermore, we optimized RF classifier by adjusting the parameter of the number of estimators to get more accurate predictions.

\subsection{Transformers}
Given the bidirectional feature of Bidirectional Encoder Representations from Transformers (BERT), BERT can understand and predict the meaning of sentences based on a Mask Language Model (MLM) technique under complex contexts. In our case, analyzing the tweets content and detecting their conspiracy features, BERT is a practical Language Model that we can leverage on in fake news detection tasks. For the purpose of assessing the performance of the model on provided datasets, we decided to experiment with BERT both with and without pretrained weights. We still first splitted our provided dataset into 80\% training set and 20\% validating set, tokenized the tweets contents from BERT tokenizer, and assigned targets with the label columns. Then, we trained the BERT model along with a two fully connected hidden layers that process the BERT hidden outputs to the prediction probabilities, and optimized the model through cross entropy loss.

\subsection{Ensemble}
In this section, we present two different ensemble methods. 

The first method is multi-layer ensemble. On the basis of the fully connected neural network classifier, we tried to use the output of different layers of Roberta to get the classification result. On the basis of the fully connected neural network classifier, we try to use the output of different layers of Roberta to get the classification result. Since different layers of a neural network can capture different levels of syntactic and semantic information, to adapt Roberta to a specific downstream task, we directly obtain result from lower layers which contain more general information. We concatenate multi-layers’ output as joint output to get the experimental results by assigning weights to the results of different layers.

The second method is multi-models ensemble. After combining multiple pre-training models, the size of the weight matrix was doubled, and a mixed output was obtained to jointly predict the class of data. This method effectively guarantees the robustness of the model. Specifically, we selected the pre-trained models Roberta and BERT, which had the same model depth, and concatenated their output embeddings into one feature matrix. When the input data passes through the pipeline, we will get the joint classification of Roberta and BERT, which can perfectly eliminate error case of one model of two, so as to get more accurate and stable results.

\section{Results and Analysis}
In this section, we present our models' Matthews Correlation Coefficient (MCC) on each of the three tasks. We also provide a thorough analysis on the experiments we conducted and motivations behind choosing the models and hyperparameters.
\begin{table}[hbt!]
  \caption{Validation MCC}
  \label{tab:graphval}
  \begin{tabular}{cccc}
    \toprule
    Model & Task 1 & Task 2 & Task 3 \\
    \midrule
    SVM & 0.422 & 0.308 & 0.205 \\
    Random Forest & 0.418 & 0.264 & 0.146\\
    BERT (w/o pretraining) & 0.339 & 0.532 & 0.436\\
    BERT (w/ pretraining) & 0.479 & 0.580 & 0.525\\
    Ensemble(multi-layers) & 0.505 & 0.553 & 0.478\\
    Ensemble(multi-models) & 0.529 & 0.567 & 0.397\\
  \bottomrule
\end{tabular}
\end{table}
\subsection{TF-IDF}
After training several models by different methods, such as naïve Bayes, decision tree, Random Forest (RF) and Support Vector Machine (SVM), we found that a RF classifier and a SVM classifier performed much better than other models. As for RF model, we fine-tuned the hyperparameter of the number of trees in the forest by trying out a wide range of values and found the number of 150 estimator reached the peak results. The best result shows that the training accuracy is 63.5\% for Task 1, 92.1\% and 89.6\% for Task 2 and Task 3 respectively. As for SVM model, the best result shows the same accuracy compared to the RF model and approximately 92.5\% for Task 2 and 89.9\% for Task 3, which performed slightly better than the RF model. Table 1 showed that the MCC score of RF and SVM models for single-label was much higher than multi-label classification, since we selected the most simple-to-implement approaches with simple additions to the initial preprocessed data.
\subsection{Transformers}
The initial results showed a training accuracy around 40\% and a testing accuracy no greater than 35\% indicating the model was not generalizing across training epochs. Therefore, we tuned the model by adjusting the hyperparameters such as hidden layer size, number of hidden layers, learning rate and number of training iterations. The best model yields approximately 61.5\% validating accuracy on text-based dataset, approximately 91.5\% accuracy on structure-based dataset, and 89.7\% accuracy on structure and text-based dataset under 516 hidden layer size, 4 hidden layers, 1e-4 learning rate and 9 epochs. Such a boost in validating accuracy after shrinking layer size and increasing learning rate is caused by the limited size of the training set and a model without pretrained weights. Typically, a model without pretrained weights needs to be trained under more epochs, smaller layer size and number of layers, and larger learning rates. Larger learning rates result in rapid changes and smaller number of layers with smaller layer size force the model converge more quickly. Those adjusted hyperparameters enables the model to find the optimal weights.
\subsection{Results of Ensemble Model}
In the multi-layer ensemble experiment, we test each hidden layer of Roberta model as an independent output. From the results, we can see that the output results of layers 11, 10 and 8 are better than the original layer's output. Therefore, we connect these layers with the original output layer to obtain the multi-layer ensemble model. The experimental results of multi-layer combined output are better than single-layer output.

In the multi-model ensemble experiment, we merge Roberta and BERT model to the ensemble model. In this experiment, we expand the model's size from 768 to 1536 and obtain the higher test accuracy in 10 epoch. We preserve the Roberta's hyperparameters of finetuning, and achieve 0.529 MCC and 69.19\%  accuracy in Task 1, achieve 0.567 MCC and 91.94\% accuracy in Task 2, much higher than both BERT and Roberta. From the training results of final round, it can be seen that Roberta has a significant influence on multi-models ensemble.




%
%

\section{Discussion and Outlook}
We present several methods to detect COVID-19 conspiracy theories from social media content. A pre-trained transformer is shown to achieve the best performance, but a transformer trained on the small dataset from scratch also yields reasonable accuracies. Our future work includes finding a better initialization than random for training an attention model from scratch as well as a more robust ensemble model for such a dataset with trending vocabularies.

\bibliographystyle{ACM-Reference-Format}
\def\bibfont{\small} 
\bibliography{sigproc} 

\end{document}